# Constraint compiling into rules formalism for dynamic CSPs computing

## Sylvain Piechowiak[1], Joaquin Rodriguez[2]

[1] LAMIH - UMR CNRS 8530 –
Université de Valenciennes - 59313 Valenciennes Cédex 9– France
email : sylvain.piechowiak@univ-valenciennes.fr
[2] INRETS-ESTAS
20, rue E. Reclus, 59650 Villeneuve d'Ascq - France
email : joaquin.rodriguez@inrets.fr

**Abstract**: *In this paper we present a rule based formalism for filtering variables domains of constraints. This formalism is well adapted for solving dynamic CSP. We take diagnosis as an instance problem to illustrate the use of these rules. A diagnosis problem is seen like finding all the minimal sets of constraints to be relaxed in the constraint network that models the device to be diagnosed.*

## 1 Introduction

Constraint programming is a representation paradigm currently extremely popular in various fields and especially in computer science. The objective targeted is to provide powerful programming tools which enable the programmers to concentrate their efforts first and foremost on the modelling of a problem rather than on its solution. It is a matter of giving priority to the "what" in comparison to the "how". Constraint programming languages are generally well-adapted to specific classes of problems. We can quote Chip [DIN, 87], Prolog III [COL, 90], Ilog Solver [PUG, 94], Oz [SMO, 95] and all the CLP(X) solvers in which X represents a specific field (integer, boolean, value interval, finite domain, etc.) [DIA, 95] [BEN, 94]. These languages prove to be more efficient for problems which can be formalized by a fixed set of constraints. Unfortunately, in reality, many problems are of a dynamic nature. In decision problems it is difficult to define exactly the constraints to be checked in order to make a decision. It is therefore necessary to be able to modify these constraints. For example in timetable generating, the constraints associated to a teacher may change if his availability varies. In on-line scheduling problems, when a machine breaks down, this modifies all the constraints of the problem: it is necessary to reschedule.

This modification can be carried out with the addition, the relaxation or the substitution of one or several constraints. Generally, constraint programming languages take these modifications into account by dealing with the resulting problem as a new problem, independent of the original one. In over-constrained problems, which are frequent in the field of decision-making aid, this way of proceeding is not acceptable. Indeed, the procedure has two major drawbacks. Firstly, there is no clear vision of the relationship which exists between one specific constraint and the problem in its entirety. Therefore, for an over-constrained problem in which the constraints cannot all be checked, it is very difficult to determine which constraints are to be relaxed. Secondly, after each modification, the same processes are performed several times, which may become very heavy if the problems are large ones.

In order to best limit the consequences of the modification of a given problem, various techniques have been defined, in particular in the community of dynamic constraint satisfaction problems [DEC, 88]. In general, they are based on systems to memorize the inferences performed in order to know at any time the link which exists between one specific constraint and the solutions being sought. These systems are quite close to the truth maintenance systems in the knowledge based systems such as the TMS [DOY, 97] or the ATMS [deK, 86].

In this paper, we propose a memorization system adapted to dynamic CSPs. This system is based on the justification of the inferences performed during the resolution of a problem. We will start therefore, in section 2, by recalling the definitions relating to CSPs. Then, in section 3, we will deal with the case of dynamic CSPs. In section 4, we will define a specific technique for constraint compiling. Finally in section 5, in order to illustrate the use of constraints compiled using this technique, we will take the example of diagnosis problem which we will treat as a dynamic CSP. We will conclude by indicating some perspectives envisaged which would improve the results obtained.

## 2    Constraint Satisfaction Problems (CSP)

### 2.1    Definitions

A static CSP or simply CSP (Constraint Satisfaction Problem) is defined by a couple $(V, C)$. $V$ is a finite set of variables $\{V_1, \ldots, V_n\}$. Each variable $V_i$ can take its values in its domain $dom(V_i)$ = $\{\gamma_{i1}, \ldots, \gamma_{iq}\}$, known beforehand. $C$ is a finite set of constraints, or relations, $\{C_1, \ldots, C_m\}$ on the variables of $V$ [MAC, 77]. The arity of a constraint is the number of variables which it connects. Generally, binary CSPs are distinguished from n-ary CSPs. The majority of research presented in the existing literature concerns binary CSPs. The first reason for this is that binary CSPs have a graph representation which makes it possible to take advantage of the numerous results known in the graph theory. The second reason is that an n-ary CSP can theoretically be transformed into an equivalent binary CSP [BAC, 98].

The graph associated to a CSP is also called a constraint network [SUS, 80] [BOO, 91]. In a graph associated to a binary CSP, the nodes represent the variables and the arcs represent the constraints. When the constraints are n-ary, the variables and the constraints are represented by different nodes. In this case, the arcs between the nodes no longer represent the constraints, they indicate a relationship of the type "is concerned by the constraint" or "concerns the variable".

The description of the domain of variables and constraints can be done in two ways: by extension or by comprehension. The description by extension consists in enumerating in an exhaustive manner all the values that each variable of the CSP can take and all the n-uples of possible values for each constraint. This representation is convenient to handle with search algorithms but it also has major drawbacks. Firstly the exhaustive enumeration is a long task as well as being at the root of errors, especially if the domains of the variables have many values. Secondly, it is very difficult to obtain a synthetic vision of a CSP described in this way.

The instantiation of a variable $V_i$ consists in associating to it one of the values taken in its domain. This will be noted : $V_i \leftarrow \gamma_{ik}$, with $\gamma_{ik} \in dom(V_i)$. A solution for a CSP corresponds to an instantiation of all the variables of this CSP in such a way so that none of the constraints are violated, that is to say that the values associated to all the variables are compatible with all of the constraints.

In a general way, a constraint satisfaction problem is expressed by a set of constraints on a set of variables. Several aims can be targeted during the processing of a CSP: to prove the existence of a solution? to find one of these solutions? Should all the solutions be defined, or characterized or counted? Should we check that an instantiation of the variables is a solution?

### 2.2    Computing CSPs

Traditionally, in the processing of a CSP, a filtering phase and an enumerating phase can be

distinguished[HAR, 80].

The filtering phase consists in removing from the domains the values which are not present in any solutions. This phase corresponds to a reduction in the search space. The skill in the filtering phase for the resolution of CSPs is to find techniques which sufficiently reduce the domains but which do not bring about excessive calculation times. The CSP obtained after the filtering phase has a degree of consistency : we speak of node consistency, arc consistency, or path consistency. Arc consistency may be obtained in reasonable time limits by algorithms such as AC3 [MAC, 77], or AC4 [MOH, 86]. Path consistency, which is better from the filtering point of view, is rarely sought because it requires extremely long calculation periods [TSA, 93].

The enumerating phase consists in instantiating  each variable of the CSP with one of the values of its domain in such a way so that all the constraints are checked. The algorithms which perform the enumeration generally proceed by progressively instantiating all the variables. As soon as the current instantiation of the variables violates one of the constraints, one of the instantiations performed is questioned. The difficulty lies in the choice of the instantiation to be questioned. This choice may simply concern the last instantiation as is done with BT (chronological backtracking). It can be more subtle as is done by BJ (back jumping) [GAS, 77], CBJ (conflict directed back jumping) [PRO, 93], etc. or various combinations of these techniques [PRO, 93b].

In the majority of research, when the CSP has no solution, the algorithms finish by merely notifying the absence of a solution. In many situations, this is not satisfactory: it is necessary to seek the origin of this absence of solution. Moreover, we are often led to modify a CSP in order to take new characteristics of a problem into account. In this case, it is desirable to be able to limit the consequences of the modifications in the search for solutions. These problems are dealt with in the dynamic CSPs.

## 3   Dynamic CSPs

In this section, we will revise the definitions relating to dynamic CSPs.

### 3.1   Basic definitions

We will say that a CSP is over-constrained when all of its constraints cannot be satisfied simultaneously, that is to say when there is no solution. There are two reasons why a CSP may be over-constrained. The first is that the domains of variables do not have enough values. In this case, in order for the CSP to have solutions, it is necessary to add values to the domains of variables. This situation corresponds, for example, to the concrete case of the search for correct sizing of the parameters of a device. The second reason is that certain constraints are incompatible with each other: in this case, in order for the CSP to have solutions, it is necessary to modify the constraints. This situation corresponds to the example of the diagnosis which we will present later in this paper.

In order to transform an over-constrained CSP into a CSP with solutions, it is necessary to modify one or more constraints. In this paper, the only transformation permitted is one which consists in removing a constraint from the set of constraints of the CSP. It should be noted that it is possible to envisage finer modifications in which one constraint could be substituted by another.

In fact, we will define a dynamic CSP as a CSP in which it is possible to add or remove one or more constraints [DEC, 88] [BES, 92] [VER, 94].

### 3.2   Propagation trace

In static CSPs with solutions there is no particular reason to explain each domain filtering process; what is important is to obtain correctly filtered domains and to enumerate the solutions.

But when the CSPs have no solution, or in the case of dynamic CSPs, it is important to know the exact origin of a value. For each value it is necessary to be able to give the set of constraints which have been used to calculate it. We must be capable of **justifying** the presence or the absence of the domains' values. Indeed, knowing these justifications makes it possible to limit propagation following each addition or removal of constraint [JUS, 96] [JUS, 97]. It also makes it possible to determine whether they are the constraints which over-constrain a CSP.

A "conflict" can be defined as a set of constraints which cannot all be satisfied at the same time. This corresponds to the notion of conflict in the ATMS [deK, 87]. The presence of a conflict is translated by the presence of a variable with an empty domain.

Generally, the algorithms for the processing of dynamic CSPs are based on the extremely delicate management of dependencies between the variables and the constraints which concern them [DEB, 95]. This is what is done by the algorithms DnAC4 [BES, 94], AC│DC [NEV, 94], DnAC6 [DEB, 94], Dynamic Backtracking [GIN, 93].

The dependency between the variables and the constraints is a simple means which makes it possible to find the origin of the conflict. However, with this means, the origin calculated is not precise. Indeed, certain constraints can, in an overall manner, connect several variables, although for certain values of these variables, the constraints do not concern all of these variables. For example, with the constraint $C_{and} : V_3 = V_1 \wedge V_2$, the knowledge that $V_1$ = false is sufficient in order to deduce that the value $V_3$ = false. On an overall level, the constraint $C_{and}$ links the variables $V_1$, $V_2$ and $V_3$, but we can be more precise on the dependencies since $C_{and}$ connects $V_3$ = false with $V_1$ = false, independently of the value of $V_2$.

In order to determine the precise origin of a conflict, it is thus necessary to define a management, not only between the variables, but also between the instantiations of the variables and the constraint propagations used to determine these instantiations. This management is all the more important in terms of processing time gain, the gain being proportionate to the number of relaxations and restoring of constraints.

The problem which now arises is to know how to exploit a constraint in order to propagate the known values of the variables in a CSP. For example, from the constraint $C_{and}(V_1, V_2, V_3) = \{(i,j,k), i \in \{true, false\}, j \in \{true, false\}, k \in \{true, false\} / k = i \wedge j\}$, according to the known values of variables, the constraint will be exploited differently by:

| |
|---|
| a1 :    if $|V1|=1 \vee |V2|=1$ then $V_3 := (V_1 \wedge V_2)$ |
| { if the domain of V1 or V2 has one value, |
|   then the domain of V3 can be reduced} |
| a2 :    if $V_3$ = true then $V_1 :=$true; $V_2 :=$true |
| a3 :    if $V_3$=false $\wedge$ $V_2$=true then $V_1 :=$false |
| a4 :    if $V_3$ =false $\wedge$ $V_1$=true then $V_2 :=$false |

Fig 1 : Propagation rules associated to $C_{and}$
|X| means domain cardinality of the variable X

The semantics of the action $V := f(V_1, V_2, \ldots, V_n)$ is :

$$\mathbf{dom(V) \leftarrow dom(V) \cap \{f(V_1, V_2, \ldots, V_n)\}}$$

This action corresponds to a domain reduction.

It is said that the values of the variables are propagated if one of the previous actions a1, a2, a3 or a4 is performed. The set of all the actions associated to a single constraint which enables its propagation is called the set of propagating procedures of the constraint.

In order to memorize the "trace" of the propagation performed, a simple technique consists in associating, to each domain deduced by propagation, the set of domains of the variables occurring in the action expression an the condition expression. This "trace will then make it possible to manage the withdrawal of certain constraints efficiently.

The trace is all the more accurate as it is limited solely to the values actually used during the propagation. Thus we could associate another set of propagating procedures to constraint $C_{and}$:

| | |
|---|---|
| a11 : | if $V_1$=true then $V_3 := V_2$ |
| a12 : | if $V_2$=true then $V_3 := V_1$ |
| a13 : | if $V_1$=false then $V_3 :=$false |
| a14 : | if $V_2$=false then $V_3 :=$false |
| a2 : | if $V_3$=true then $V_1 :=$true; $V_2 :=$true |
| a3 : | if $V_3$=false $\wedge$ $V_2$=true then $V_1 :=$false |
| a4 : | if $V_3$=false $\wedge$ $V_1$=true then $V_2 :=$false |

Fig 2: Other propagating rules associated to $C_{and}$

When the variable $V_1$ or $V_2$ gets the value "false", in the two cases (Fig 1 & 2), the propagation will lead to the deduction of the false value for $V_3$. However, in the rules a13 et a14 of fig2 the trace obtained links the false value of $V_3$ with just one of the two instantiations ($V_1\leftarrow$false) or ($V_2\leftarrow$false), unlike the rule a1 of fig1 in which the origin of the instantiation of $V_3$ cannot be precisely determined. The quality of the reasoning carried out using a CSP is directly affected by the fineness of the management of dependency links in this same CSP. The following example (Fig 3) illustrates this remark.

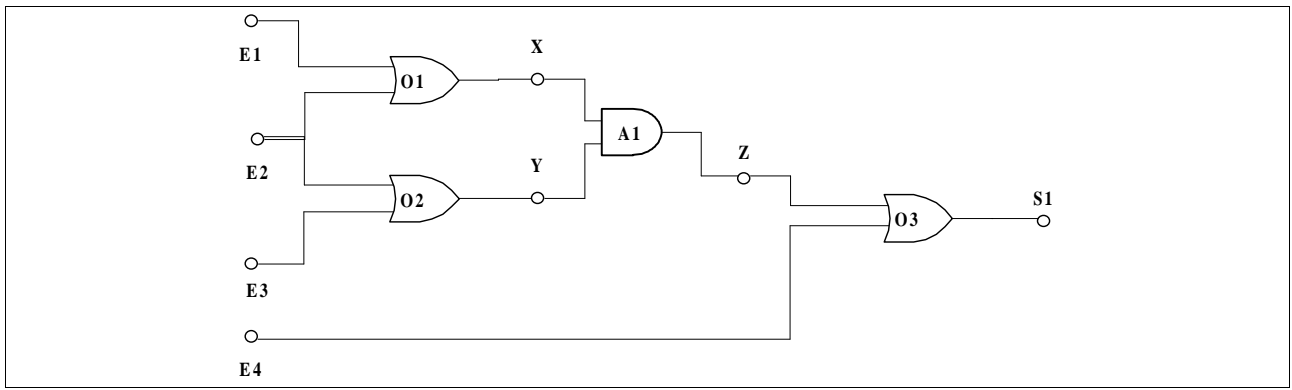

Fig 3: Example of a constraint network

Let us suppose that we have E1=E2=E3=S1=false and E4=true. O1, O2 and O3 are $C_{or}$ constraints and A1 is a $C_{and}$ constraint. Initially, the propagating procedures associated to $C_{and}$ and $C_{or}$ are respectively {a1, a2, a3, a4} and {b1, b2, b3, b4}.

| | |
|---|---|
| b11 : | if $V_1$=false then $V_3 :=V_2$ |
| b12 : | if $V_2$=false then $V_3 :=V_1$ |
| b13 : | if $V_1$=true then $V_3 :=$true |
| b14 : | if $V_2$=true then $V_3 :=$true |
| b2 : | if $V_3$=false then $V_1 :=$false; $V_2 :=$false |
| b3 : | if $V_3$=true $\wedge$ $V_2$=false then $V_1 :=$true |
| b4 : | if $V_3$=true $\wedge$ $V_1$=false then $V_2 :=$true |

Fig 4: Propagating rules associated to $C_{or}$

From E1=false, E2=false, it can be deduced that X=false by propagation on O1. From E2=false, E3=false, it can be deduced that Y=false by propagation on O2. From X=false, Y=false, it can be deduced that Z=false by propagation on A1. And finally, from E4=true, Z=false, it can be deduced that S1=true by propagation on O3. A conflict can be seen on S1: {O1,O2,A1,O3}.

Let us now associate the set of rules of propagation to O1, O2 and O3:

This time, from E4=true, by propagation on O3, we are immediately led to S1=true. The conflict {O3} appears more rapidly. If we seek to explain this conflict, in the first case we would have to relax successively O1, O2, O3 and A1 to conclude that O3 is responsible for the conflict whereas this result is obtained directly in the second case.

## 4    Constraint propagation rules

In [PIE, 00] we have proposed an algorithm called GENERATE to find a set of rules associated with a boolean constraint on n variables. Briefly, to find a rule R like :

$$\text{IF } (V_1 = x_1) \text{ and } (V_2 = x_2) \text{ THEN } V_3 := x_3 \; V_3 := x_4$$

the GENERATE algorithm looks in the truth table if no line states that

$$(V_1 = x_1) \text{ and } (V_2 = x_2) \text{ and } (V_3 <> x_3) \text{ or that } (V_1 = x_1) \text{ and } (V_2 = x_2) \text{ and } (V_3 <> x_4)$$

The algorithm begins with the rules with smaller set of conditions.

Here, we only remember the properties checked using the rules obtained by this algorithm.

### 4.1    Properties of the rules obtained by the GENERATE procedure

**(cr1) semantic equivalence of the representations**: all the local propagation made in one of the representations must be done with the other and vice-versa and no solution is lost when filtering with the propagation rules representation.

 **(cr2) correctness of the rules**: if several rules are potentially usable, their successive firing use must not lead to a contradiction concerning the variables. When the rules check the cr1 property and contradictions appear, this means that the constraint itself can not be satisfied.

**(cr3) independence of the rules**: if several rules are potentially usable, the choice of one of them to propagate the constraint does not matters.

 **(cr4) rule minimality** : The condition part of the rules must be the most restrictive possible and the conclusion part must deduce the most information on variables. In a general way, if we have two rules, R1 and R2, so that condition (R1) = condition(R2)$\wedge\xi$, only rule R1 is retained. If conclusion(R1) includes conclusion(R2) then the rule R2 is retained.

Example: given the logical constraint $C_{and}(V_1,V_2,V_3)$ : $V_3 = (E1 \wedge E2)$. The set of propagating rules associated to $C_{and}$ obtained is:

> R1 :    IF $(V_1$ = true) and $(V_2$ = true) THEN $V_3$ := true
> R2 :    IF $(V_1$ = false) THEN $V_3$ := false
> R3 :    IF $(V_2$ = false) THEN $V_3$ := false
> R4 :    IF $(V_3$ = true) THEN $(V_1$ :=true; $V_2$ := true)
> R5 :    IF $(V_1$ = true) and $(V_3$ = false) THEN $V_2$ := false
> R6 :    IF $(V_2$ = true) and $(V_3$ = false) THEN $V_1$ := false

### 4.2    Exploitation of the rules

Thanks to the GENERATE procedure, each constraint is represented by a set of propagating rules which check the properties cr1 to cr4. During the propagation of values in a network of constraints expressed with this type of form, a "pattern-matching" process is sufficient to choose the type of propagation of the constraint. In the case where the domains of variables are enumerated, after the use of one of the rules associated to a constraint, it is not necessary to consider the other rules associated with this constraint.

```
Procedure propagate (in C: a constraint)
begin
choose R*∈ C so that conditions (R*)=true
if R* does not exist
        then C cannot be propagated
        else for each rule R from C do mark R as propagated endfForEach
"perform" conclusions(R*)
associate R* to each modification
endIf
end
```

The conclusions( R) part of each rule is a set of actions like V := f(V1,...,Vn). Performing a such conclusions part consists in reducing the domain of the variable V according to the expression f(V1,...,Vn).

The relaxation of a constraint C is accompanied by the cancellation of all the propagation which start from C: this is what the procedure **relax** does. The re-establishment of a C constraint which had previously been relaxed is accompanied by the restoration of all the propagation which starts from C: this is what the procedure **restore** does.

Marking does not bring about destruction but rather a masking of the values and activating of rules. In this way, when we wish to cancel the relaxation of a constraint, only some update work on the masking is necessary: no propagation is necessary. This management of relaxation by masking is not given in detail here. Let us simply underline the fact that it limits propagation in a network of constraints, which is of prime importance when many relaxation procedures are performed.

```
Procedure relax(in C :a constraint)
begin
for each rule R of C do
    cancelRule(R)
    mark R as cancelled
endfForEach
end
```

```
Procedure restore(inC :a constraint)
begin
For each rule R of C do
    restoreRule(R)
    remove cancellation marker from R
endfForEach
end
```

```
Procedure cancelRule(in R :a rule)
begin
for each CONC of conclusion(R) do
    cancelCondition(CONC)
    add a cancellation marker to CONC
endfForEach
end
```

```
Procedure restoreRule(in R :a rule)
begin
for each conclusionCONC of R do
    restoreCondition(CONC)
    remove cancellation marker from CONC
endfForEach
end
```

```
Procedure cancelCondition (in COND: a conclusion)
begin
for each rule R which has COND in condition do
    cancelRule(R)
endfForEach
end
```

```
Procedure restoreCondition(in COND : a conclusion)
begin
for each rule R which has COND in condition do
    restoreRule(R)
endfForEach
end
```

Let us remark, finally, that it is possible to manage the markers in a simple way, by using a support counter as is done in the algorithms DnAC4 and DnAC6. This management can also be carried out by indexing according to various information. In [ROD, 94], the authors present an indexing system according to time.

## 5 Application to Constraint-based reasoning

Constraint-based reasoning has many applications in decision problems like in device diagnosis research. In short, the idea pursued is to model a diagnostic problem as a dynamic CSP and the aim is to determine which constraints must be relaxed in order for the CSP to allow solutions.

Here we take the problem of the diagnosis as an example of constraint-based reasoning in which we apply the resolution rules generated by the GENERATE algorithm.

According to [VER, 99], in decision tools, it is not sufficient to conclude "no solution" when a CSP has no solution. The user of these tools want to be explained why the CSP has no solution. One way consists in find the maximal sets of constraints with some solutions. In other words: explain why a CSP has no solution can be explained by finding the minimal sets of constraints to relax to restore consistency of the constraint network. In model-based diagnosis community a such explanation is called a diagnosis.

Let us revise rapidly what is meant by diagnosis. The correct functioning of a device is modelled by a CSP. Certain variables are instantiated: they correspond to the observations made on the real device. When the device is faulty, the CSP becomes over-constrained and the goal is to find explanations or diagnosis.

The search for the minimal sets of constraints to be removed in order to restore consistency requires the accurate knowledge of the origin, or of the justification, of the values of the variables. The representation of the constraints by sets of resolution rules facilitates the management of dependencies between values.

We will explain the reasoning method using the example the circuit in Fig 6.

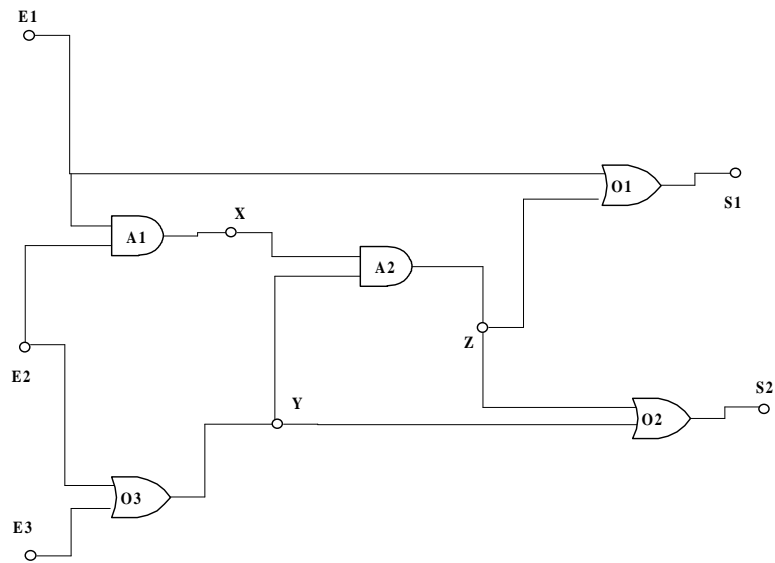

Fig 6: CIRC1: a device to be diagnosed

Graphically (Fig 7), each constraint is represented by a set of "variable" nodes associated to the variables, a set of "rule" nodes representing the propagating rules and by links between the "variable" nodes and the "rule" nodes.

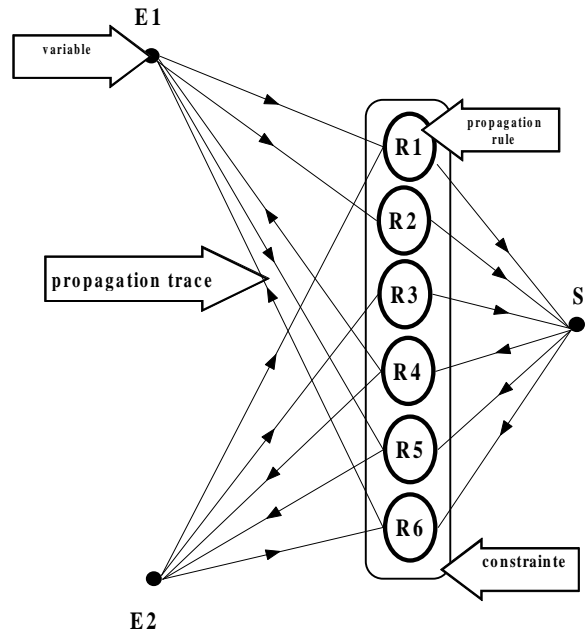

Fig 7: Graphical representation of a constraint and variables

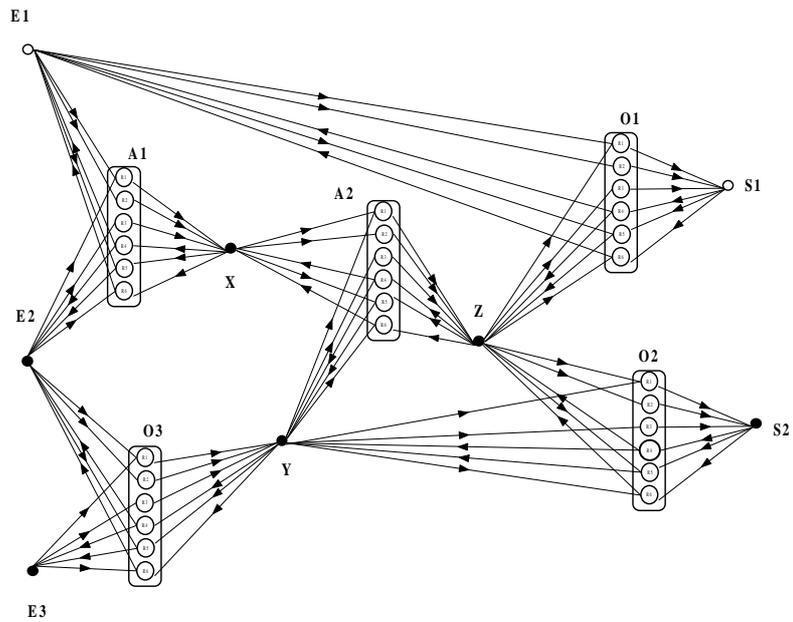

Fig 8: the CSP associated to CIRC1

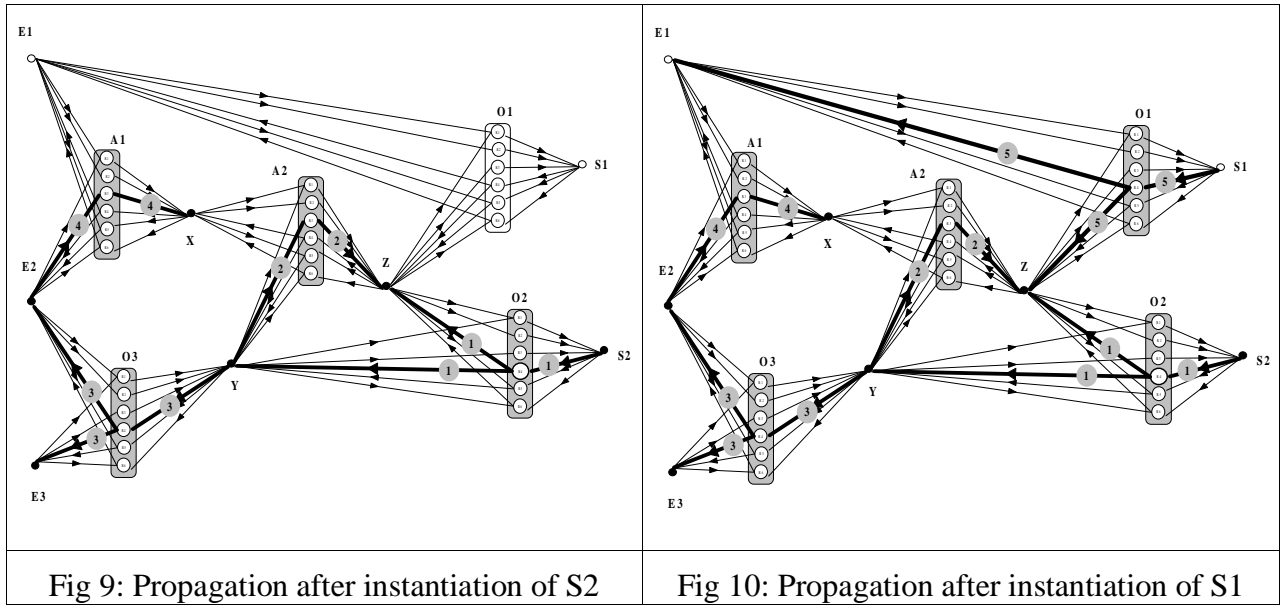

| Fig 9: Propagation after instantiation of S2 | Fig 10: Propagation after instantiation of S1 |

The initial domain of the variables is BOOL={true,false}.

**Step 0** : Before any affectation of variables, their domain includes all of its possible values. No rule has been activated for these values. This comes down to justifying the values of the domain by emptiness. In our example, as all the variables are boolean, we associate the justification →{true,false}to them.

**Step 1**: The variable S2 is instantiated to false. This value is propagated as is shown with bold arrows of propagation trace in Fig 9. The fired rules are numbered with grey circles.

**Step 2** : The variable S1 is instantiated to false, which brings about the propagation shown in Fig 10. After the instantiations of S1 and S2, and solely with these two instantiations, the domain of all the variables is reduced to one single value: the CSP thus allows one single solution.

**Step 3** : Let us suppose that we now instantiate the variable E2 to true. The domain of E2 is reduced to {} : the CSP becomes over-constrained and no longer has a solution. The origin of this over-constraint can be explained using the conflict {O3, O2}. In the following chart, we show the evolution of the propagations during the various steps.

A possible modification of the domain of a variable corresponds to each propagation. The value deduced by this propagation is justified by the use of a rule of the constraint considered. This rule is associated to the value deduced. For example, in the chart, during the second step, variable E1 is modified by rule R4 of constraint O1. This makes it possible to deduce the false value. This link is memorized by R4/O1→{false}.

In order to determine which constraints should be relaxed so that the CSP has solutions, the starting point is the conflict obtained after all the propagations {O3, O2}.

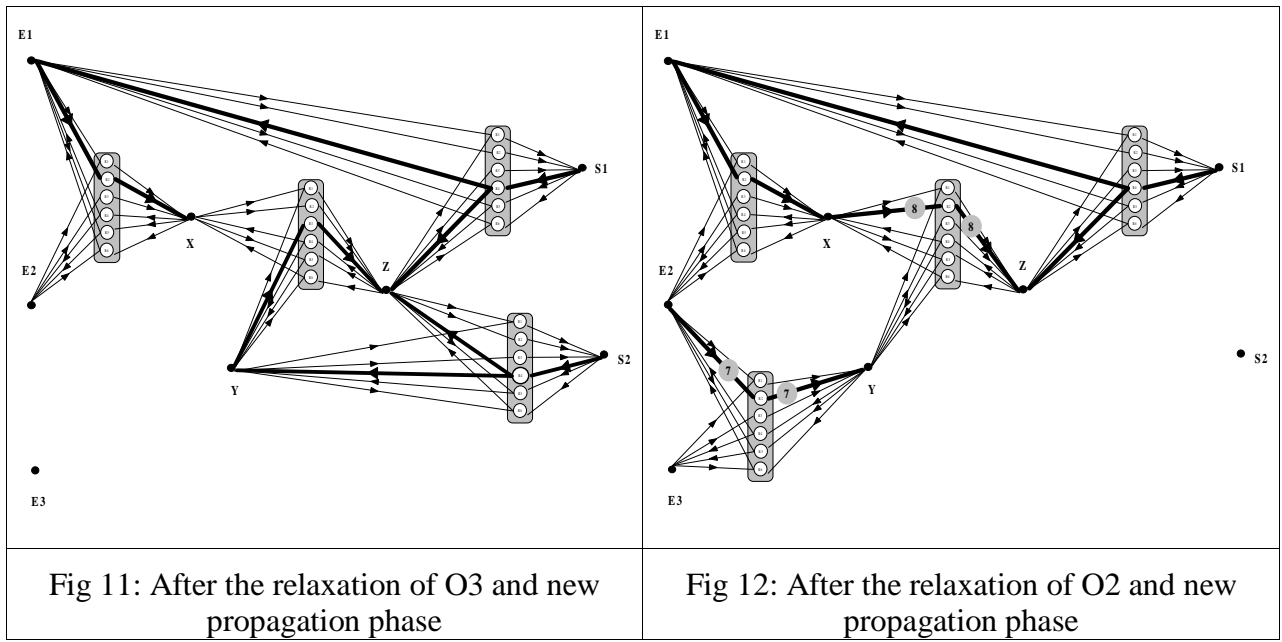

| Fig 11: After the relaxation of O3 and new propagation phase | Fig 12: After the relaxation of O2 and new propagation phase |

The relaxation of O3 or O2, along with a new propagation phase makes the conflict disappear: there are thus two possible explanations for the conflict: {O3} and {O2}. In terms of diagnosis, these explanations are diagnoses. Each relaxation is illustrated by Fig 11 & 12. After each relaxation performed, the new CSP allows solutions.

The ATMS make it possible to explain conflicts but on the condition that these conflicts are minimal [deK, 87]. Here we do not present this hypothesis, which forces us to enter a new phase of relaxation if conflict is not minimal. The only hypothesis we make is that if a conflict exists, it will show itself in one of the steps.

## 6 Comparison with other works

Little research has been carried out up to now in the field of constraint compiling for dynamic CSPs. There are links with the research performed in the field of reasoning maintenance systems of the TMS or ATMS type. However, the monotony constraint which enables the ATMS to work in an incremental manner is no longer valid here. The role of an ATMS is to justify the inferences carried out by a system which infers. When a conflict appears, the ATMS provides explanations for the inference system. These explanations are only valid to the extent that the facts used during the inference are not questioned.

The representation of a constraint in the form of a set of solving rules is close to the representation in the form of a set of functions which is presented in [SAN, 93]. However, the authors do not consider the question of dynamic CSPs. The management of dependencies is therefore dealt with.

In [DAV, 93] and [LIU, 95], the authors show interesting properties in a functional representation. A functional representation consists in expressing a constraint by a set of functions which calculate the value of one of the variables using the others. Unfortunately, a functional representation is sometimes likely to lead to a loss of information. To illustrate this statement, let us return to the case of the constraint : S = (E1 ∧ E2). When the value of S is true, we can deduce that E1= true and E2=true. This deduction cannot be expressed by functional representation.

In [APT, 99] the RULES GENERATION gives rules that have the format: X = s → y <> a where X is a subset of variables from C, s are values from the domains of X, y is a variable from C not in X and a a value from the domain of y. This algorithm ensure a rule consistency that is a weaker notion of local consistency. In this paper author extend their algorithm to an other that generate inclusion rules. This algorithm ensure arc-consistency.

In [ABD, 00] an other algorithm called RuleMiner is proposed. This one gives rules like the Generate algorithm proposed in [PIE, 00] for boolean domain. In addition, RuleMiner can also process finite domain.

## 7  Conclusion, Extensions and Perspectives

We have presented a specific technique for the compiling of constraints which is well adapted for the processing of CSPs. It consists in transforming a constraint into a set of equivalent propagation rules. The properties checked by the rules associated to a single constraint make it possible to  achieve efficient processing of the dynamic CSPs, especially when they are over-constrained. This technique leads to the problem of the generation of solving rules. In this paper, we have suggested a constraint representation with propagation rules. This algorithm deals with the constraints expressed in the form of truth tables. Finally, we have illustrated the use of this representation of constraints in a diagnosis search problem according to the model-based approach.

One of the perspectives directly linked  to the research presented in this paper concerns the processing of other types of value domains : enumerated or value intervals [HYV, 88] [DAV, 87].

The principle of using the solving rules to represent the constraints is valid for boolean constraints, but is also valid for constraints in various domains. On the other hand, the generation of these rules is difficult to compute. This is one of the research lines we are currently following. We are using this representation to take into account the temporal dimension in the field of diagnosis. Very briefly, we have associated a date (or a time interval) to each value handled. The principle of resolution rules enabled us to manage a dependency not only between values and constraint, but also a temporal dependency. The main difficulty we have met appeared when the time intervals associated to the values of the variables which appear in the condition part were different to those associated to the values of the conclusion variables.